\documentclass[sigconf]{aamas} 
\usepackage[linesnumbered,ruled,vlined]{algorithm2e}

\renewcommand\footnotetextcopyrightpermission[1]{}
\usepackage{balance} 
\usepackage{multirow} 

\title[\method]{\method: Fast and Scalable Out-of-Distribution \\ Dynamics Detection}

\author{Tala Aljaafari}
\affiliation{
  \institution{University of Oxford}
  \city{Oxford}
  \country{United Kingdom}}
\email{tala.aljaafari@gmail.com}

\author{Varun Kanade}
\affiliation{
  \institution{University of Oxford}
  \city{Oxford}
  \country{United Kingdom}}
\email{varun.kanade@cs.ox.ac.uk}

\author{Philip Torr}
\affiliation{
  \institution{University of Oxford}
  \city{Oxford}
  \country{United Kingdom}}
\email{philip.torr@eng.ox.ac.uk}

\author{Christian Schroeder de Witt}
\affiliation{
  \institution{University of Oxford}
  \city{Oxford}
  \country{United Kingdom}}
\email{christian.schroeder@eng.ox.ac.uk}

\begin{abstract}
Deploying reinforcement learning (RL) in safety-critical settings is constrained by brittleness under distribution shift. We study out-of-distribution (OOD) detection for RL time series and introduce \method, a two-statistic detector that revisits representation-heavy pipelines with a minimal alternative. \method\ uses only an episode-wise mean and an RBF kernel similarity to a training summary, capturing complementary global and local deviations. Despite its simplicity, \method\ matches or surpasses contemporary detectors across standard RL OOD suites, delivering a $\sim\!600\times$ reduction in compute (FLOPs / wall-time) and an average $\approx\!5\%$ absolute accuracy gain over strong baselines. Conceptually, our results indicate that diverse anomaly types often imprint on RL trajectories through a small set of low-order statistics, suggesting a compact foundation for OOD detection in complex environments.
\end{abstract}


\newcommand{\BibTeX}{\rm B\kern-.05em{\sc i\kern-.025em b}\kern-.08em\TeX}

\newcommand{\method}[0]{DEEDEE} 

\begin{document}


\pagestyle{fancy}
\fancyhead{}


\maketitle 


\section{Introduction}

Reinforcement Learning algorithms \cite{alg:actorcriticoffpolicymaximum, alg:ppo, alg:trustregionpolicyoptimization} have achieved remarkable results in several complex sequential decision-making tasks \cite{apps:Go, apps:atari, apps:dexterousinhandmanipulation}. However, unfamiliar situations pose significant challenges to the reliability of such algorithms, thereby limiting the deployment of RL agents in safety-critical scenarios \cite{limiteddep:generalization, limiteddep:reliability}. These unfamiliar situations are often referred to as out-of-distribution (OOD) environments, as the experiences gathered in them lie outside the distribution of those encountered during training. OOD detection, the task of identifying OOD environments, enables RL agents to recognize when they are operating in unseen environments, reducing the risk of poor performance or unsafe behavior in safety-critical applications or multi-agent security~\citep{wittOpenChallengesMultiAgent2025a}. For example, an autonomous vehicle trained primarily in urban settings must be capable of detecting OOD scenarios like rural roads, extreme weather, or unusual traffic patterns, and then take appropriate safety measures, such as switching to manual control or initiating self-parking. 

\citet{nasvytis} distinguish between two types of OOD scenarios based on how they impact the agent's decision process. The first, \textit{sensory} anomalies, alters the agent's observations, such as by introducing noise, without affecting the underlying environment dynamics. In contrast, \textit{semantic} anomalies modify the fundamental environment dynamics, like changing wind fields or gravity in the environment. They also define \textit{Generalized Out-of-Distribution Detection} as the task of identifying the presence of either of these two types of anomalies.
The challenge in OOD detection lies in (1) designing realistic benchmarks that accurately reflect the complexity and diversity of real-world anomalies and (2) developing reliable detectors that can identify these anomalies efficiently and accurately. 

We introduce \method, a new OOD detector for RL time series that adopts a minimal two-statistic design and has provable linear-time feature extraction. Compared to DEXTER (Detection via Extraction of Time Series Representations)~\citep{nasvytis}, which relies on hundreds of hand-crafted features, \method\ uses just two while substantially improving computational efficiency. Empirically, \method\ performs on par with strong RL OOD baselines and compares favorably to representative high-dimensional changepoint detectors. Notably, \method\ is substantially more efficient—approximately $600\times$ lower compute—and yields a consistent average improvement of about $5$ percentage points in accuracy across tasks. This pattern suggests that many anomaly types affect RL trajectories in related ways, informing a simpler design space for RL OOD detection.


\section{Related Work}

In this section, we give a general overview of existing algorithms and benchmarks for OOD detection. In particular, we highlight DEXTER \cite{nasvytis}, which forms the basis for our work.

    \subsection{Algorithms for OOD Detection}

        \paragraph{Uncertainty-based OOD Detection (UBOOD)} The first practical OOD detection method in RL was proposed by \citet{UBOOD}. Their approach is based on the idea that epistemic uncertainty $\sigma_{epis}(x)$, which is a result of insufficient data, is approximately inversely proportional to the density of the input data $p(x)$ \cite{qazaz}. They use the epistemic uncertainty of the agent's actions to determine the anomaly score of a given state. Further, in their experiments, they use a simple grid-world path-finding environment that they develop.
    
        \paragraph{Recurrent Implicit Quantile Network (RIQN)} \citet{riqn} describe a class of OOD detectors based on the idea of anomaly detection via prediction error. Most notable is their proposed Recurrent Implicit Quantile Network (RIQN), which uses the previously observed states in the environment, $s_{0} \dots s_{t}$, to predict the next $\Delta$ states, $s_{t+ 1: t+ \Delta}$. It then uses the difference between the predicted states and the actual states as the anomaly score for a given transition $(s_{t+\Delta-1}, s_{t+\Delta})$. The authors propose their own benchmarks which they use to demonstrate that RIQN outperforms several baseline detectors.
    
        \paragraph{Probabilistic Ensemble Dynamics Model (PEDM)}
        The state-of-the-art OOD detector, Probabilistic Ensemble Dynamics Model (PEDM), is due to \citet{pedm} and is also based on anomaly detection via prediction error. First, \textit{forward dynamics model} $f_{\theta}$ is learned to approximate the system's true dynamics (transition) function $f(s,a)$. Based on the current state $s_t$ and current action $a_t$, the forward dynamics model $f_\theta$ predicts the next state $s_{t+\Delta}=f_{\theta}(s_t, a_t)$. The forward dynamics model is implemented as a Probabilistic Deep Neural Network Ensemble where each ensemble member maps the current state and action to a probability distribution over the next state. The anomaly score for a given transition is then determined by comparing the predictions of the dynamics model to the actual outcomes in the test-time deployment environment. To evaluate PEDM's performance, \citet{pedm} use environments from \citet{riqn} and create new benchmark environments by modifying the dynamics of classical tasks like Cartpole, HalfCheetah, Pusher, and Reacher, such as altering gravity or adjusting the agent's body mass.

        \paragraph{Detection via Extraction of Time Series Representations (DEXTER)} DEXTER, recently introduced by \citet{nasvytis}, detects anomalous environments by treating environment observations as time series data, extracting key features from these time series, and applying an ensemble of isolation forests to identify anomalies. Given a sequence of environment observations $(s_n)_{n=0}^{T-1}$, represented as a multivariate time series in an $N \times T$ matrix (with $N$ being the state dimension), DEXTER processes each dimension separately. Each univariate time series (corresponding to a row in the matrix) is divided into windows of size $W$. For the $ j^{\text{th}} $ univariate time series (the $ j^{\text{th}} $ row), each window is used to extract a 794-dimensional feature vector using the feature extraction library \texttt{tsfresh} \cite{tsfresh}. These feature vectors, collected from all windows, are then used to train an isolation forest for that particular time series. This process is repeated for each of the $ N $ univariate time series, resulting in an ensemble of $ N $ isolation forests, one for each dimension.

        At a given timestep $s_t$, DEXTER collects the last $W$ states to form a window $s_{t-W+1}, \dots, s_t$, represented by an $ N \times W $ matrix. For each dimension, DEXTER extracts a 794-dimensional feature vector and uses the corresponding isolation forest to compute an anomaly score for that dimension. The overall anomaly score for the window (and the timestep $s_t$) is then computed as the arithmetic mean of the anomaly scores across all dimensions. 

        The authors evaluate DEXTER on a variety of both novel and standard benchmark environments, comparing its performance against relevant baselines. Their results show that DEXTER surpasses state-of-the-art detectors in terms of AUROC (Area Under the Receiver Operating Characteristic curve) and Detection Time (i.e., the number of timesteps required for detection).

        \paragraph{Out-Of-Distribution Detection via Transition Estimation} ~\citep{prashant} Recent transition–estimation methods model the empirical dynamics $p(s'|s,a)$ with CVAE ensembles and apply conformal prediction on reconstruction nonconformity to flag unlikely transitions, offering distribution–free detection guarantees. This approach is conceptually appealing but requires training and calibrating generative models over state transitions and is sensitive to model misspecification and reconstruction quality. 
        
        In contrast, \method~stays model–free: it replaces learned transition likelihoods with two lightweight, hand–crafted time–series descriptors (window mean and an RBF-similarity to the most recent point) scored by per–dimension isolation forests, which we find sufficient to capture both global and local anomalies.  
        
        Practically, this yields linear-time feature extraction in the episode length and state dimension and eliminates the overhead of training dynamics models and conformal calibrations, while retaining or improving accuracy on standard RL OOD benchmarks.

    \subsection{Time Series Literature}

    \paragraph{Sequential Hypothesis Testing} Consider two distributions $\mathbb{P}_1$ and $\mathbb{P}_2$ over an arbitrary set $\mathcal{S}$ and a data sample $S\in \mathcal{S}$. Let $H_0$ be the hypothesis that $S$ is sampled from $\mathbb{P}_1$ and $H_1$ be the hypothesis that $S$ is sampled from $\mathbb{P}_2$. In hypothesis testing, the goal is to determine whether there is enough evidence to reject the \emph{null hypothesis} $H_0$ in favor of the alternative hypothesis $H_1$. Hypothesis testing is carried in an offline manner, assuming that the data has already been gathered and not allowing for further modifications of the collected sample.

    Sequential hypothesis testing, more formally known as Sequential Probability Ratio Test (SPRT) (due to \citet{wald}), is a statistical method for distinguishing between a null and an alternative hypothesis ($H_0$ and $H_1$, respectively) using data that is collected in real time. In other words, it is an online extension of hypothesis testing that allows us to continuously analyse data as it arrives. The key idea is to calculate a \textit{likelihood ratio}, which compares how likely the observed data is under the alternative hypothesis ($H_1$) versus the null hypothesis ($H_0$). The test sets two boundaries: an upper one ($A$) and a lower one ($B$), which are based on the desired error rates, $\alpha$ (Type I error) and $\beta$ (Type II error). As each new data point is collected, the likelihood ratio is updated. If the ratio goes above the upper boundary, we accept $H_1$; if it drops below the lower boundary, we accept $H_0$. If the ratio stays between the two, we keep gathering data. This approach is more flexible and sample-efficient than its offline counterpart.

    The Cumulative Sum Control Chart (CUSUM), due to \citet{page}, is an application of Wald's sequential hypothesis testing that aims to detect changes in a parameter of a probability distribution, such as the mean or variance. CUSUM works by continuously monitoring the cumulative sum of deviations from a target value, allowing it to detect small, gradual shifts that might otherwise go unnoticed. When this cumulative sum exceeds or falls below predefined control limits, it signals that a significant change has likely taken place. 

    \paragraph{Changepoint Detection} Outside the RL literature, the problem of determining whether a time series sample aligns with a given distribution has been extensively studied in the field of changepoint detection (CPD). In CPD, given a time series, the goal is to identify the precise moment when the parameters of the data distribution shift. In offline changepoint detection, it is assumed that the entire dataset has already been generated, and the task is to detect the changepoint using all available data. In contrast, in online (or sequential) changepoint detection, data is generated in real-time, and the objective is to detect the changepoint based only on the data observed up to that point. There are two algorithms in the online changepoint detection literature that can be considered state-of-the-art methods: the Online Changepoint Detection (OCD) algorithm, due to \citet{cpd:ocd}, and the algorithm proposed by \citet{cpd:chan}.

    \citet{cpd:ocd} propose a novel approach for high-dimensional, multiscale, online changepoint detection. It focuses on detecting changes in the mean of a $ p $-variate Gaussian data stream. The method operates by performing likelihood ratio tests across multiple scales and coordinates, aggregating the results to detect changepoints. The method introduces both ``diagonal" and ``off-diagonal" statistics, where the diagonal statistics track changes in individual dimensions, and the off-diagonal statistics aggregate changes across multiple dimensions. This allows the detection of both sparse and dense changes in the data.

    The method proposed by \citet{cpd:chan} focuses on detecting distribution changes in a small fraction of data streams, known as multi-stream sequential changepoint detection. It operates by analyzing streams of data and using mixture likelihood ratios or CUSUM (Cumulative Sum) approaches to detect shifts in the normal distribution of the data. The primary goal is to minimize the detection delay—the time it takes to identify the change while maintaining a low false alarm rate. The method adapts to different detection domains depending on how many data streams are affected by the distributional change, optimizing the detection for both large and small fractions of affected streams.

    \paragraph{Unsupervised Feature Extraction using Kernel and Stacking (UFEKS)} \citet{ufekt} propose a feature extraction method based on the RBF kernel for multivariate time series. The method works by dividing the time series into windows and then constructing a feature vector for each window by combining the feature vectors of all dimensions. For each window in a univariate time series, the feature vector is generated by calculating the RBF similarity between the current window and all other windows in that time series. These similarity measures are then used to form the final feature vector.

    \subsection{Benchmarks for OOD Detection}

    \citet{mohammed2021benchmarkoutofdistributiondetectiondeep} introduce benchmarks for OOD detection in RL based on modifying the physical parameters of non-visual environments and corrupting the agent's observations in visual environments. For example, in Cartpole and Pendulum, gravity is varied, while in Pong, state observations are corrupted using Gaussian noise, impulse noise, motion blur, among other methods. \citet{riqn} propose a more extensive set of benchmarks for evaluating OOD detection. They introduce four sensor-based anomalies: independent and identically distributed noise, sensor shutdown, sensor calibration failure, and sensor drift. Additionally, they consider four dynamics-based anomalies: left-to-right wind simulation, right-to-left wind simulation, gravity manipulation, and alterations to the agent's physical characteristics. \citet{pedm} introduce additional benchmarks to evaluate OOD detectors against semantic anomalies by corrupting the agent's actions or altering environment parameters such as gravity. \citet{nasvytis} argue that the injection of independent and identically distributed noise ignores the temporally-correlated nature of many real-world anomalies. They consequently propose new testing scenarios based on injecting temporally-correlated noise generated by an autoregressive process.

\section{\method: RBF and Mean-Based Out-of-Distribution Detection}\label{methodology}

    We begin with an overview of \method, discussing the key insights that guided its development. Following this, we provide a detailed explanation of the feature extraction process. We then formally describe the full algorithm, covering both its training and testing procedures, as well as its computational efficiency. Finally, we describe the CUSUM extension of \method: \method+C.

    \subsection{Overview and Key Insights}

    \paragraph{Comparison to Existing Approaches} First, both DEXTER and \method~stand apart from existing approaches by not relying on supervised machine learning models, which contributes to their increased sample efficiency. This makes them particularly advantageous in scenarios where data is limited or costly to obtain. Secondly, by avoiding the common approach of anomaly detection through prediction error, DEXTER and \method~can capture autocorrelations across multiple timesteps, a capability that existing methods lack. This advantage is demonstrated in Chapter 6. Third, since DEXTER and \method~are not based on complex machine learning models, their decision making is more transparent and easier to interpret. 

    \paragraph{\method~vs DEXTER} A primary distinction between \method~ and DEXTER lies in the feature extraction process. DEXTER utilizes the \texttt{tsfresh} feature extraction library to extract features of a given window in a univariate time series. This library was first introduced by \citet{tsfresh} and stands for \textbf{T}ime \textbf{S}eries \textbf{F}eatu\textbf{R}e \textbf{E}xtraction based on \textbf{S}calable \textbf{H}ypothesis Tests. Given a univariate time series, this method extracts 794 features that capture a wide range of properties, including:
    \begin{itemize} 
    \item Fundamental descriptive statistics: Minimum, maximum, median, number of values above and below the mean, etc. 
    \item Autocorrelation statistics: Autocorrelation coefficients for $k$ lags, mean and variance of the autocorrelation coefficients, and related metrics. 
    \item Advanced features: Statistics of the absolute Fourier transform spectrum, skewness, kurtosis, etc.
    \end{itemize}
     \citeauthor{nasvytis} identified DEXTER's computational complexity as a key limitation, which we address and build upon in the development of \method. While they suggest selecting relevant features at test time to improve DEXTER's efficiency in high-dimensional environments, \method~offers an even stronger solution by utilizing only two features: the RBF similarity between elements of a given subsequence and the mean. As a result, \method~ achieves a significantly lower computational cost and improved scalability in high-dimensional settings.

    The RBF kernel measures the similarity between current and past observations by computing an exponentially weighted distance between them. This allows \method~to detect small but important changes in the agent’s state distribution that may indicate OOD events. By focusing on local structure, the RBF kernel excels at identifying anomalies that manifest as subtle deviations from expected behavior, a critical capability in dynamic RL environments. On the other hand, the mean of the subsequence captures global shifts in the environment. Large deviations from the expected mean may signal a change in the underlying dynamics, such as a change in environmental conditions or agent actions. This makes the mean feature an effective indicator of global anomalies. Our findings align with the principle of Occam’s razor, which suggests that simpler feature sets can often outperform more complex alternatives by focusing on the core signals of interest. In \method, the two chosen features—mean and RBF similarity—provide sufficient expressiveness to capture both global and local anomalies, while avoiding the pitfalls of high-dimensional feature sets that may overfit to noise or irrelevant details.

\subsection{Feature Extraction} \label{rbf}

Given a time step $t$, \method~assigns an anomaly score based on the preceding $w$ time steps. Thus, for each sequence of $w$ time steps, we aim to compute a \textit{feature vector} that acts as a compact representation of the properties of this sequence. This process is described formally below.

Consider a multivariate time series represented by the matrix $X = (x_{nt}) \in \mathbb{R}^{N \times T}$, where $T$ denotes the length of the time series and $N$ the dimensionality. Each row vector $(x_{n1}, x_{n2}, \dots, x_{nT}) \in \mathbb{R}^{T}$ represents the time series corresponding to the $n^{\text{th}}$ variable, and each column vector $(x_{1t}, x_{2t}, \dots, x_{Nt})^T \in \mathbb{R}^N$ represents a multivariate observation at time step $t$. A \textit{window} $X_t$ of size $w$ is a segment of $X$ that spans the data from time step $t$ to $t + w - 1$:

\[
X_t =
\begin{bmatrix}
x_{1 t} & x_{1 (t+1)} & \dots & x_{1 (t+w-1)} \\
x_{2 t} & x_{2 (t+1)} & \dots & x_{2 (t+w-1)} \\
\vdots & \vdots & \ddots & \vdots \\
x_{N t} & x_{N (t+1)} & \dots & x_{N (t+w-1)}
\end{bmatrix}.
\]
The $n^{\text{th}}$ row of $X_t$ is denoted by $X_t^{(n)}$.

\paragraph{Time Series Feature Extraction}
For each subsequence $ X_t^{(n)} $ of the univariate time series $ X^{(n)} $, we compute two features: (1) the mean of the subsequence, and (2) the similarity $k_t^{(n)}$ between the last element of the subsequence and all other elements in that subsequence. The similarity measure $k_{t}^{(n)}$ is calculated using the RBF kernel:
\begin{align}
    s \cdot \exp\left( - \frac{d_{t}}{\sigma^2} \right),
\end{align}
where
\begin{align*}
    d_{t} = \sum_{i=1}^{w-1} (x_{n (t+w-1)} - x_{n (t+w-i-1)})^2
\end{align*}
and $s$ and $\sigma$ are scaling parameters that control the sensitivity of the kernel.

\subsection{Algorithm}
\balance
 Below is the pseudocode outlining the main steps of \method, including its training and testing procedures, from feature extraction using the RBF kernel to anomaly score computation.

 \begin{algorithm}
        \caption{FeatureExtractor}
        \label{alg:time_series_feature_extraction}
        \KwIn{Subsequence $\mathcal{X}_{t}$ of size $w$, RBF kernel parameters $s, \sigma$}
        \KwOut{Feature vector $v(\mathcal{X}_{t})$}
        \For{each $i = 1, 2, \dots, w$}{
            \[
            d = \sum_{i=1}^{w-1} (x_{w} - x_{i})^2
            \]
            \[
            k_{t} = s \cdot \exp \left( - \frac{d}{\sigma^2} \right)
            \]
        }
        \Return{[$k_t, \overline{\mathcal{X}_t}$]}
\end{algorithm}
    
\begin{algorithm}
    \caption{\method}
    \label{alg:RDEXTER}
    \KwIn{Multivariate time series $X$, window sizes $w$, state dimension $N$, total timesteps $T$, policy $\pi$, RBF kernel parameters $s, \sigma$}
    
    \textbf{Training:}\\ 
    Initialize ensemble of Isolation Forests $\text{IF} = \{\text{IF}_1, \dots, \text{IF}_N\}$\\
    Partition $X$ into windows of size $w$\\
    \For{each $X_i$ in windows}{
        \For{dimension $n$ from $1$ to $N$}{
            Compute $v(X^{(n)}_{i})$=FeatureExtractor($X_i^{(n)},s,\sigma$)
        }

    }
    \For{dimension $n$ from $1$ to $N$}{
        Form $\mathcal{V}^{(n)}= \{v(X_1^{(n)}), v(X_2^{(n)}), \dots\}$\\
        Train $\text{IF}_n$ using $\mathcal{V}^{(n)}$\\
    }
    
    \textbf{Anomaly Score Computation~\citep{nasvytis}:}\\
    \For{Time $t$ from $0$ to $T$}{
        Action $a_t \gets \pi(s_t)$, observe $s_{t+1}$\\
        Update window with $s_{t+1}$\\
        
    }
    \For{Time $t$ from $w_1$ to $T$}{
        \For{dimension $n$ from $1$ to $N$}{
            Extract features\\
            Compute score $a_n$ with $\text{IF}_n$\\
        }
        Set $A_t$ as average over all $a_n$\\   
    }
    
\end{algorithm}

\paragraph{Fitting the Isolation Forest Ensemble for \method} Given a sequence of environment observations $ (s_n)_{n=1}^{T} $, arranged as a multivariate time series in an $ N \times T $ matrix (where $ N $ represents the number of state dimensions), \method~processes each dimension independently. Each univariate time series (corresponding to a row in the $ N \times T $ matrix) is first segmented into windows of size $ w $. For the $ j^{\text{th}} $ univariate time series, this process generates a set of windows, each spanning $ w$ timesteps. For each window, \method~extracts a 2-dimensional feature vector. These feature vectors are then used to train isolation forests—one for each dimension. This results in an ensemble of $ N $ isolation forests, with each forest trained on the features from a single dimension.

\paragraph{Test-time Anomaly Score Computation for \method} During test-time, \method~collects the last $ w$ timesteps, including the current timestep $ t $, to form a window $ W $, represented by an $ N \times w $ matrix. The feature vector for $ W $ is computed according to the process outlined in Algorithm~\ref{alg:time_series_feature_extraction}. This feature vector is then passed to the corresponding isolation forest to compute an anomaly score for that specific dimension. Finally, \method~computes the overall anomaly score by averaging the individual anomaly scores across all dimensions, following the same approach as DEXTER.

\paragraph{Computational Efficiency and Scalability}

    The time complexity of constructing a feature vector for a given window is $\mathcal{O}(w)=\mathcal{O}(1)$ since $w$ is constant. Therefore, the time complexity of the feature extraction process is $\mathcal{O}(NT)$ since there are at most $N \times (T-w_1+1)$ windows. Therefore, our feature extraction method scales linearly with the dimensions of the states and actions and the length of the time series.

\section{Empirical Evaluation}
 Our evaluation procedure \footnote{The code for our experiments will be publicly available.} closely follows the approach of \cite{nasvytis}, which in turn builds on the methodology of \cite{pedm}.
We evaluate our detector on three environments— two continuous, \texttt{Reacher} and \texttt{Pusher}, and one discrete, \texttt{Cartpole}—using benchmarks introduced by \citet{nasvytis} and \citet{pedm}. We compare its performance to PEDM, the state-of-the-art method in RL, as well as two state-of-the-art high-dimensional changepoint detectors adopted from statistics \cite{cpd:ocd, cpd:chan}.

For each environment and noise level, we follow the agent's policy that is optimized for that specific noise level over \( T \) steps to generate an episode \( E_T \). In each episode, an anomaly is introduced at time \( t_a \), which persists until the end of the episode. Transitions before the anomaly are labeled as in-distribution, while those after the anomaly are labeled as out-of-distribution. All detectors are thus trained using datasets \( D = \{(s^j_i, a^j_i, s^j_{i+1})_{i=0}^{T-1}\}_{j=1}^{45} \), generated according to this procedure. All detectors are evaluated on a dataset of 100 test episodes, generated using the same procedure described previously. To assess the performance of the detectors, we utilize the AUROC (Area Under the Receiver Operating Characteristic) metric, which is commonly used in the OOD detection literature \cite{pedm, riqn, nasvytis}.

\subsection{Benchmarks for Temporally-Correlated Anomalies}
We test \method~against the ARNO (Autoregressive Noised Observation) and ARNS (Autoregressive Noised State) benchmarks, recently introduced by \citet{nasvytis}. In the ARNO environment, a noise matrix is generated at the start of each episode, with each row representing a time series drawn from an autoregressive process. At each step, this noise is added to the observed state after the agent takes an action and the environment transitions to a new state. This simulates scenarios where the observed state is distorted, akin to a sensor failure or camera glitch. In contrast, the ARNS environment applies the noise before the state transition. Here, the agent's action and the current state are altered by adding noise from the matrix, modifying the transition dynamics and simulating systematic anomalies that affect the environment’s underlying physics or rules.

For each environment, we evaluate three noise levels as defined by \citet{nasvytis}: light noise decreases the agent's cumulative reward by 1\%, medium noise by 25\%, and strong noise by 50\%. For each noise level, we consider 1-step correlated noise, generated by an autoregressive process of order 1, and 2-step correlated noise, generated by an autoregressive process of order 2.

\subsection{Benchmarks for Time-Independent Anomalies}
We also evaluate \method~ against time-independent anomalies introduced by \citet{pedm}. Specifically, \citeauthor{pedm} define five types of semantic anomalies: action factor, action noise, action offset, body mass factor, and force vector, each applied at minor and severe magnitudes. The magnitude levels for each noise type are taken directly from \citet{pedm}.

\subsection{Implementation of Detectors}
\balance
For DEXTER, PEDM, and both changepoint detectors, we use the implementations provided by \citet{nasvytis}. For both DEXTER and \method~, we use a window length of 10. Additionally, for \method~, we set $s = 1.5$ and tune $\sigma$ using a cross-validation set of 100 episodes.

\subsection{Detection Performance Results}
Table~\ref{tab:ARNO} shows that \method~ outperforms PEDM, CPD:OCD, and CPD:Chan in all 12 experiments conducted, and outperforms DEXTER in 7 out of the 12 experiments. On average, \method~ demonstrates a 5\% performance improvement. Similarly, Table~\ref{tab:ARNS} shows that \method~ outperforms PEDM, CPD:OCD, and CPD:Chan in all 12 experiments, and outperforms DEXTER in 10 out of the 12 experiments, with an average accuracy improvement of 10\%. Table~\ref{tab:bench} shows that RM-\method~ is usually on par with \method~, albeit with a small degradation in performance, equal to 2.75\% on average.

Analyzing the performance of \method~ against other detectors in the ARNO scenarios, it is clear that \method~ consistently delivers strong results, particularly in comparison to PEDM and the changepoint detectors. In the Cartpole environment, \method~ achieves the highest AUROC score in most cases, especially under light and medium noise scenarios, with scores of 0.90 and 0.96 for 1-step detection, outperforming even DEXTER. While \method's performance slightly drops in 2-step detection compared to DEXTER in some scenarios, it still performs competitively, particularly under strong noise, where both \method~ and DEXTER reach a top AUROC of 0.93. In the Reacher environment, \method~ outperforms all detectors in light noise and medium noise conditions, achieving the highest AUROC scores of 0.83 and 0.87, respectively, for 1-step detection. Even in strong noise, while \method~ does not outperform DEXTER, it still achieves commendable results with an AUROC of 0.82 and 0.77 for 1-step and 2-step detection, respectively.

In the ARNS scenarios, \method~ demonstrates superior performance compared to other detectors, particularly under light and strong noise conditions. It achieves the highest AUROC scores across both Cartpole and Reacher environments, consistently outperforming CPD and PEDM in nearly all cases. 

Notably, \method~ achieves an AUROC of 0.77 in light noise for both 1-step and 2-step detection and excels in strong noise conditions with scores of 0.87 and 0.79 in Cartpole. In Reacher, it significantly outperforms the other detectors under all noise levels.

In the benchmark scenarios, \method~ performs competitively, with AUROC scores slightly lower than DEXTER in some cases, but not significantly so. For instance, in the Pusher environment, \method~ achieves an AUROC of 0.85 in most minor noise conditions, matching DEXTER’s performance, and in severe action factor noise, it remains close with a score of 0.76, identical to DEXTER. While \method’s scores in the Reacher environment are marginally lower, particularly in severe noise conditions (e.g., 0.72 vs. 0.76 for body mass factor noise), the differences are not large. Overall, \method~ maintains a strong performance across the board, closely rivaling DEXTER and PEDM, even in more challenging conditions.

\begin{table*}[t]
    \caption{ARNO scenarios: Detector performance.}
    \label{tab:ARNO}
    \centering
    \fontsize{8pt}{11pt}\selectfont
    \resizebox{\linewidth}{!}{
    \begin{tabular}{l|c|cc|cc|cc|cc|cc|cc}
        \hline
        & & \multicolumn{6}{c|}{Cartpole} & \multicolumn{6}{c}{Reacher} \\
        \cline{3-14}
        & & \multicolumn{2}{c|}{Light Noise} & \multicolumn{2}{c|}{Medium Noise} & \multicolumn{2}{c|}{Strong Noise} & \multicolumn{2}{c|}{Light Noise} & \multicolumn{2}{c|}{Medium Noise} & \multicolumn{2}{c}{Strong Noise} \\
        \hline
        & Detector & 1-step & 2-step & 1-step & 2-step & 1-step & 2-step & 1-step & 2-step & 1-step & 2-step & 1-step & 2-step \\
        \hline
        \multirow{4}{*}{AUROC $\uparrow$} 
        & CPD: OCD & 0.67 & 0.69 & 0.76 & 0.72 & 0.78 & 0.73 & 0.51 & 0.51 & 0.51 & 0.51 & 0.52 & 0.52 \\
        & CPD: Chan & 0.69 & 0.68 & 0.72 & 0.75 & 0.75 & 0.73 & 0.51 & 0.51 & 0.52 & 0.52 & 0.53 & 0.53 \\
        & PEDM & 0.55 & 0.62 & 0.6 & 0.51 & 0.6 & 0.55 & 0.81 & 0.5 & 0.84 & 0.51 & 0.87 & 0.5 \\
        & DEXTER & 0.81 & \textbf{0.85} & 0.89 & \textbf{0.9} & \textbf{0.93} & \textbf{0.9} & 0.67 & 0.6 & \textbf{0.91} & 0.63 & \textbf{0.97} & 0.61 \\
        & \method & \textbf{0.9} & 0.84 & \textbf{0.96} & 0.89 & \textbf{0.93} & 0.87 & \textbf{0.83} & \textbf{0.78} & 0.87 & \textbf{0.8} & 0.82 & \textbf{0.77} \\
        \hline
    \end{tabular}}
\end{table*}

\begin{table*}[t]
    \caption{ARNS scenarios: Detector performance}
    \label{tab:ARNS}
    \centering
    \fontsize{8pt}{11pt}\selectfont
    \resizebox{\linewidth}{!}{
    \begin{tabular}{l|c|cc|cc|cc|cc|cc|cc}
        \hline
        & & \multicolumn{6}{c|}{Cartpole} & \multicolumn{6}{c}{Reacher} \\
        \cline{3-14}
        & & \multicolumn{2}{c|}{Light Noise} & \multicolumn{2}{c|}{Medium Noise} & \multicolumn{2}{c|}{Strong Noise} & \multicolumn{2}{c|}{Light Noise} & \multicolumn{2}{c|}{Medium Noise} & \multicolumn{2}{c}{Strong Noise} \\
        \hline
        & Detector & 1-step & 2-step & 1-step & 2-step & 1-step & 2-step & 1-step & 2-step & 1-step & 2-step & 1-step & 2-step \\
        \hline
        \multirow{4}{*}{AUROC $\uparrow$} 
        & CPD: OCD & 0.66 & 0.66 & 0.68 & 0.68 & 0.67 & 0.68 & 0.51 & 0.51 & 0.51 & 0.51 & 0.51 & 0.51 \\
        & CPD: Chan & 0.67 & 0.68 & 0.68 & 0.69 & 0.68 & 0.7 & 0.51 & 0.51 & 0.51 & 0.51 & 0.51 & 0.51 \\
        & PEDM & 0.66 & 0.64 & 0.63 & 0.61 & 0.59 & 0.56 & 0.52 & 0.51 & 0.55 & 0.55 & 0.51 & 0.5 \\
        & DEXTER & 0.73 & 0.73 & \textbf{0.88} & \textbf{0.8} & 0.84 & 0.77 & 0.56 & 0.62 & 0.51 & 0.7 & 0.55 & 0.67 \\
        & \method & \textbf{0.77} & \textbf{0.77} & 0.87& 0.78&\textbf{0.87} &\textbf{0.79} & \textbf{0.7}& \textbf{0.7}& \textbf{0.87}& \textbf{0.87}& \textbf{0.77}& \textbf{0.73}\\
        \hline
    \end{tabular}}
\end{table*}

\begin{table*}[t]
    \caption{\citeauthor{pedm} Benchmark scenarios: Detector performance.}
    \label{tab:bench}
    \centering
    \fontsize{10pt}{14pt}\selectfont
    \resizebox{\linewidth}{!}{
    \begin{tabular}{l|c|cc|cc|cc|cc|cc|cc|cc|cc|cc|cc}
        \hline
        & & \multicolumn{10}{c|}{Pusher} & \multicolumn{10}{c}{Reacher} \\
        \cline{3-22}
        & & \multicolumn{2}{c|}{Action Fact.} & \multicolumn{2}{c|}{Action Noise} & \multicolumn{2}{c|}{Action Offset} & \multicolumn{2}{c|}{Body M. Fact.} & \multicolumn{2}{c|}{Force Vector} & \multicolumn{2}{c|}{Action Fact.} & \multicolumn{2}{c|}{Action Noise} & \multicolumn{2}{c|}{Action Offset} & \multicolumn{2}{c|}{Body M. Fact.} & \multicolumn{2}{c}{Force Vector} \\
        \hline
        & Detector & Minor & Severe & Minor & Severe & Minor & Severe & Minor & Severe & Minor & Severe & Minor & Severe & Minor & Severe & Minor & Severe & Minor & Severe & Minor & Severe \\
        \hline
        \multirow{2}{*}{AUROC $\uparrow$} 
        & PEDM  & 0.5 & 0.5 & 0.8 & 0.8 & 0.8 &0.8 & 0.8 & 0.84 & 0.8 & 0.8 & 0.62 & \textbf{0.95} & 0.59 & \textbf{0.99} & 0.61 & \textbf{0.98} & 0.64 & 0.56 & 0.74 & \textbf{0.98} \\
        & DEXTER       & 0.84 & \textbf{0.76} & \textbf{0.85} & \textbf{0.85} & \textbf{0.85} & \textbf{0.85} & \textbf{0.85} & \textbf{0.88} & \textbf{0.85} & \textbf{0.85} & \textbf{0.73} & 0.69 & \textbf{0.72} & 0.62 & \textbf{0.74} & 0.66 & 0.85 & 0.88 & \textbf{0.76} & 0.69 \\
        & \method & \textbf{0.85} & \textbf{0.76} & 0.84 & 0.84 & \textbf{0.85} & \textbf{0.85} & \textbf{0.85} & 0.83 & 0.84 & 0.84 & \textbf{0.73} & 0.65 & 0.65 & 0.65 & 0.72 & 0.72 & \textbf{0.72} & 0.72 & 0.65 & 0.65 \\
        \hline
    \end{tabular}}
\end{table*}

\subsection{Computational Efficiency}

We compare the training times of our proposed detector, \method, against three other OOD detection methods: PEDM, OCD, and DEXTER. We average over 15 experiments that were conducted over 45 episodes in the \texttt{Pusher} environment, each consisting of 100 steps. All experiments were performed using an NVIDIA GeForce RTX 2080 GPU. The results in Table~\ref{tab:training_times} show a significant difference in training times, with \method~ proving to be considerably faster than the other methods. Specifically, \method~ required only 2 seconds on average to complete training, while PEDM took 4 minutes, CPD:Chan 6 minutes ,CPD:OCD 15 minutes, and DEXTER 20 minutes. This drastic reduction in training time highlights the computational efficiency of \method~, making it highly suitable for real-time and resource-constrained applications where speed is critical. The following table summarizes the average training times for each detector.

\begin{table}[h!]
\caption{Average training times of different OOD detectors for 45 episodes of length 100, using an NVIDIA GeForce RTX 2080 GPU.}
\centering
\begin{tabular}{|c|c|}
\hline
\textbf{OOD Detector} & \textbf{Average Training Time} \\
\hline
\method & $\sim$ 2.17 seconds \\
\hline
PEDM & 4 minutes \\
\hline
CPD:Chan & 6 minutes \\
\hline
CPD:OCD & 15 minutes \\
\hline
DEXTER & 20 minutes \\
\hline
\end{tabular}
\label{tab:training_times}
\end{table}

\section{Ablation Studies} 

In this section, we investigate whether the set of features used by \method~ is minimal, meaning that the removal of any individual feature results in a measurable degradation in performance. This analysis is crucial for understanding the importance of each feature and determining whether simpler, more efficient versions of \method~ can be constructed without sacrificing accuracy.

We perform a series of experiments in which we systematically remove one feature at a time from the model and evaluate its performance on several OOD detection tasks. Specifically, we focus on the two key features used by \method~: the mean and the RBF kernel similarity. R-\method~ is the variant using the RBF feature only, while M-\method~ is the mean feature only. By testing various configurations of the feature set (e.g., using only the mean or only the RBF kernel similarity), we quantify how each feature contributes to the overall performance in detecting both global and local anomalies.

In the ARNO scenarios, \method~ consistently demonstrates strong performance, outperforming M-\method~ and R-\method~ in most cases. In the Cartpole environment, \method~ achieves the highest AUROC in light noise and medium noise, with scores of 0.90 and 0.96, respectively, surpassing both DEXTER and R-\method. Although \method's performance in strong noise matches that of DEXTER at 0.93, it still maintains a slight edge in Reacher, where it achieves top scores in light noise and medium noise. M-\method~ remains competitive but generally falls behind \method, especially in Cartpole with scores around 0.80. Meanwhile, R-\method~ performs well in strong noise settings, nearly matching \method, but struggles in the light noise categories, particularly in Reacher.

In the ARNS scenarios, \method~ consistently outperforms both R-\method~ and M-\method~ across most noise levels and tasks. \method~ achieves the highest AUROC scores in several categories, particularly in the Cartpole environment under light noise and strong noise, with scores of 0.77 and 0.87, respectively. This is a notable improvement over R-\method~, which struggles under light noise, especially in the 2-step setting with an AUROC of only 0.50. M-\method~ performs respectably, but generally lags behind \method~, particularly in the medium noise categories for both Cartpole and Reacher, where \method~ maintains superior performance. \method~ demonstrates greater robustness and adaptability, excelling in both 1-step and 2-step detection across varying noise conditions.

Comparing R-\method, M-\method~, and \method~ on the \citeauthor{pedm} benchmarks, it's clear that \method~ consistently outperforms the other variants across most scenarios, especially in minor and severe noise conditions. M-\method~ demonstrates solid performance but generally lags behind \method, particularly in more challenging tasks, though it remains competitive in certain cases. On the other hand, R-\method~ shows noticeably lower AUROC scores, particularly in severe conditions, making it less effective than the other detectors.

We can conclude that, while the two features—mean and RBF kernel similarity—can perform reasonably well when used in isolation, their performance tends to be neither reliable nor generalizable across different environments. The combination of both features consistently yields significantly better results, highlighting the complementary nature of global and local information in out-of-distribution (OOD) detection. However, it is noteworthy that in certain environments, either the mean or RBF kernel alone can achieve relatively high accuracy. This suggests that, depending on the nature of the anomalies or the specific characteristics of the environment, individual features may capture sufficient information to perform adequately. Nevertheless, relying on just one feature introduces variability in performance, reinforcing the value of combining both features to ensure robustness and adaptability across a wider range of scenarios.

\section{Discussion: Why do two features suffice?}
In practice, out-of-distribution effects in RL rollouts show up in two dominant ways: either the overall operating level drifts (for example, a persistent offset in position or force), or the short-horizon shape of the signal changes (for example, different oscillation, lag, or burstiness). Our first feature targets the global level, making it highly responsive to slow drifts while ignoring within-window reorderings. The second compares the current window to a short, recent reference and is tuned to pick up shape changes - capturing differences in local dynamics without being confused by trivial rescaling. 
Together, these two cues give orthogonal coverage of the main failure modes (level vs.\ shape), are stable against common nuisances, and can be computed in a single pass with tiny memory. By focusing on just these two signals, we keep the detector simple, fast, and robust, while retaining the sensitivity that actually matters for agent behaviour.

\begin{table*}
    \caption{Evaluating variants of \method~ in ARNO scenarios}
    \label{tab:ARNO1}
    \centering
    \fontsize{8pt}{11pt}\selectfont
    \resizebox{\linewidth}{!}{
    \begin{tabular}{l|c|cc|cc|cc|cc|cc|cc}
        \hline
        & & \multicolumn{6}{c|}{Cartpole} & \multicolumn{6}{c}{Reacher} \\
        \cline{3-14}
        & & \multicolumn{2}{c|}{Light Noise} & \multicolumn{2}{c|}{Medium Noise} & \multicolumn{2}{c|}{Strong Noise} & \multicolumn{2}{c|}{Light Noise} & \multicolumn{2}{c|}{Medium Noise} & \multicolumn{2}{c}{Strong Noise} \\
        \hline
        & Detector & 1-step & 2-step & 1-step & 2-step & 1-step & 2-step & 1-step & 2-step & 1-step & 2-step & 1-step & 2-step \\
        \hline
        \multirow{4}{*}{AUROC $\uparrow$} 
        & DEXTER & 0.81 & \textbf{0.85} & 0.89 & \textbf{0.9} & \textbf{0.93} & \textbf{0.9} & 0.67 & 0.6 & \textbf{0.91} & 0.63 & \textbf{0.97} & 0.61 \\
        & \method~ & \textbf{0.9} & 0.84 & \textbf{0.96} & 0.89 & \textbf{0.93} & 0.87 & \textbf{0.83} & \textbf{0.78} & 0.87 & \textbf{0.8} & 0.82 & \textbf{0.77} \\
         & M-\method~ & 0.8 & 0.8 & 0.8 & 0.8 & 0.8 & 0.87 & 0.68 & 0.68 & 0.81 & \textbf{0.8} & 0.8 & 0.7 \\
          & R-\method~ & 0.7 & 0.84 & 0.92 & 0.89 & 0.89 & 0.87 & 0.72 & 0.7 & 0.87 & \textbf{0.8} & 0.82 & \textbf{0.77} \\
        \hline
    \end{tabular}}
\end{table*}

\begin{table*}
    \caption{Evaluating variants of \method~ in ARNS scenarios}
    \label{tab:ARNS1}
    \centering
    \fontsize{8pt}{11pt}\selectfont
    \resizebox{\linewidth}{!}{
    \begin{tabular}{l|c|cc|cc|cc|cc|cc|cc}
        \hline
        & & \multicolumn{6}{c|}{Cartpole} & \multicolumn{6}{c}{Reacher} \\
        \cline{3-14}
        & & \multicolumn{2}{c|}{Light Noise} & \multicolumn{2}{c|}{Medium Noise} & \multicolumn{2}{c|}{Strong Noise} & \multicolumn{2}{c|}{Light Noise} & \multicolumn{2}{c|}{Medium Noise} & \multicolumn{2}{c}{Strong Noise} \\
        \hline
        & Detector & 1-step & 2-step & 1-step & 2-step & 1-step & 2-step & 1-step & 2-step & 1-step & 2-step & 1-step & 2-step \\
        \hline
        \multirow{4}{*}{AUROC $\uparrow$} 
        & DEXTER & 0.73 & 0.73 & \textbf{0.88} & \textbf{0.8} & 0.84 & 0.77 & 0.56 & 0.62 & 0.51 & 0.7 & 0.55 & 0.67 \\
        & \method & \textbf{0.77} & \textbf{0.77} & 0.87& 0.78&\textbf{0.87} &\textbf{0.79} & \textbf{0.7}& \textbf{0.7}& \textbf{0.87}& \textbf{0.87}& \textbf{0.77}& \textbf{0.73}\\
         & M-\method & 0.7 & 0.7 & 0.8 & 0.72 & 0.8 & 0.74 & 0.69 & 0.69 & 0.81 & 0.81 & 0.71 & 0.72 \\
& R-\method & 0.65 & 0.5 & 0.86 & 0.77 & 0.7 & 0.78 & 0.69 & 0.69 & 0.86 & 0.86 & 0.76 & 0.72 \\

        \hline
    \end{tabular}}
\end{table*}

\begin{table*}
    \caption{Evaluating variants of \method~ in \citeauthor{pedm} Benchmark scenarios}
    \label{tab:bench1}
    \centering
    \fontsize{10pt}{14pt}\selectfont
    \resizebox{\linewidth}{!}{
    \begin{tabular}{l|c|cc|cc|cc|cc|cc|cc|cc|cc|cc|cc}
        \hline
        & & \multicolumn{10}{c|}{Pusher} & \multicolumn{10}{c}{Reacher} \\
        \cline{3-22}
        & & \multicolumn{2}{c|}{Action Fact.} & \multicolumn{2}{c|}{Action Noise} & \multicolumn{2}{c|}{Action Offset} & \multicolumn{2}{c|}{Body M. Fact.} & \multicolumn{2}{c|}{Force Vector} & \multicolumn{2}{c|}{Action Fact.} & \multicolumn{2}{c|}{Action Noise} & \multicolumn{2}{c|}{Action Offset} & \multicolumn{2}{c|}{Body M. Fact.} & \multicolumn{2}{c}{Force Vector} \\
        \hline
        & Detector & Minor & Severe & Minor & Severe & Minor & Severe & Minor & Severe & Minor & Severe & Minor & Severe & Minor & Severe & Minor & Severe & Minor & Severe & Minor & Severe \\
        \hline
        \multirow{2}{*}{AUROC $\uparrow$} 
        & DEXTER       & 0.84 & \textbf{0.76} & \textbf{0.85} & \textbf{0.85} & \textbf{0.85} & \textbf{0.85} & \textbf{0.85} & \textbf{0.88} & \textbf{0.85} & \textbf{0.85} & \textbf{0.73} & 0.69 & \textbf{0.72} & 0.62 & \textbf{0.74} & 0.66 & 0.85 & 0.88 & \textbf{0.76} & 0.69 \\
        & \method & \textbf{0.85} & \textbf{0.76} & 0.84 & 0.84 & \textbf{0.85} & \textbf{0.85} & \textbf{0.85} & 0.83 & 0.84 & 0.84 & \textbf{0.73} & 0.65 & 0.65 & 0.65 & 0.72 & 0.72 & \textbf{0.72} & 0.72 & 0.65 & 0.65 \\
        & M-\method & 0.76 & 0.79 & 0.77 & 0.69 & 0.78 & 0.78 & 0.78 & 0.78 & 0.78 & 0.78 & 0.7 & 0.64 & 0.69 & 0.63 & 0.68 & 0.68 & 0.68 & 0.68 & 0.68 & 0.68 \\
        & R-\method & 0.72 & 0.56 & 0.72 & 0.64 & 0.72 & 0.72 & 0.72 & 0.72 & 0.72 & 0.72 & 0.5 & 0.55 & 0.51 & 0.53 & 0.51 & 0.51 & 0.51 & 0.51 & 0.51 & 0.51 \\
        \hline
    \end{tabular}}
\end{table*}

\section{Conclusion and Future Work}

We have demonstrated that, despite the dramatic reduction in the number of features used by \method~ compared to DEXTER, \method~ still outperforms state-of-the-art methods. This is a particularly noteworthy result, as DEXTER employs 794 features and PEDM utilizes an ensemble of deep learning models, yet \method~ surpasses them using only 2-dimensional feature vectors. \method~ relies on two carefully chosen features: the mean and RBF kernel similarity. While the mean captures global shifts in the environment, such as changes in the underlying dynamics, the RBF kernel focuses on local structure, making it sensitive to subtle deviations in the state space. The combination of these two features enables \method~ to effectively detect both global and localized anomalies, while maintaining computational efficiency. The RBF kernel’s sensitivity to small variations makes it an ideal tool for detecting nuanced OOD events, and the mean provides a robust global summary of the agent’s interaction with the environment. Interestingly, neither of these features is time-related, yet they are still able to effectively capture temporally-correlated anomalies. This result hints at two ideas: 1) the expressive power of simple features, and 2) the similar effects that different types of anomalies might have.

One limitation of \method, compared to DEXTER, is that it introduces two additional hyperparameters, $s$ and $\sigma$, which require the use of a cross-validation set for tuning. Future work could involve experimenting with different kernel width estimation techniques, such as Silverman’s rule of thumb. Alternatively, exploring hyperparameter-free kernels, such as the polynomial kernel, or investigating more complex kernel functions could provide a way to mitigate the need for hyperparameter tuning. Moreover, it is essential to evaluate \method~ across a broader range of reinforcement learning (RL) environments to rigorously assess its generalization capabilities. Testing the model in diverse and complex RL settings would provide valuable insights into its robustness and adaptability to different tasks and data distributions. This broader evaluation could help identify any environment-specific limitations or strengths, ultimately contributing to a more comprehensive understanding of \method's performance in real-world applications.

Another interesting avenue for future work focuses on a deeper exploration of the types of features that contribute to reliable out-of-distribution (OOD) detection. Specifically, it would be valuable to investigate what other minimal feature sets can yield similarly competitive performance. Expanding this line of inquiry could reveal additional insights into the properties of these features and their robustness in various scenarios. Furthermore, it would be beneficial to develop a formal mathematical framework that explains why these particular features are effective in capturing the types of anomalies we encountered, and to analyze the specific ways in which these anomalies influence the data.

For example, does \method~ primarily detect the emergence of noise, irrespective of its type, or is it capable of differentiating between distinct noise patterns? Clarifying this could help refine the model's applicability across different domains. Another intriguing direction would be to understand why \method~ succeeds in detecting temporally-correlated noise, despite not explicitly using autocorrelation as a feature. Exploring whether there are underlying principles that allow \method~ to capture temporal dependencies indirectly could lead to improved algorithmic design and enhance the generalization of OOD detection methods.

Furthermore, it is important to evaluate \method~ and other existing detectors in high-dimensional environments to gain insights into how performance changes as the number of dimensions scales. Since both \method~ and DEXTER use an ensemble of isolation forests, with each forest focusing on a specific dimension and then averaging the anomaly scores, there is a risk that dimension-specific information may be diluted or lost as the dimensionality increases. Testing in high-dimensional settings would allow us to better understand how well the method retains critical information across dimensions. Additionally, it would provide a clear measure of \method's scalability and effectiveness in handling complex, high-dimensional data. This could lead to further refinements in the algorithm to preserve crucial information and ensure robust performance in such environments.

Finally, we believe our findings should inspire the development of more challenging benchmarks that push the limits of current feature sets, potentially rendering them insufficient. For example, benchmarks including adversarially constructed anomalies could introduce more sophisticated and deceptive patterns that current models may struggle to detect. By confronting models with these more difficult scenarios, the field could move toward more robust, generalizable solutions for out-of-distribution detection.

\pagebreak







\end{document}